\documentclass{article}

\usepackage[preprint]{neurips_2023}

\usepackage[utf8]{inputenc} 
\usepackage[T1]{fontenc}    
\usepackage{hyperref}       
\usepackage{url}            
\usepackage{booktabs}       
\usepackage{amsfonts}       
\usepackage{nicefrac}       
\usepackage{microtype}      
\usepackage{xcolor}      

\usepackage[utf8]{inputenc} 
\usepackage[T1]{fontenc}    
\usepackage{hyperref}       
\usepackage{url}            
\usepackage{booktabs}       
\usepackage{amsfonts}       
\usepackage{nicefrac}       
\usepackage{microtype}      
\usepackage{xcolor}         

\usepackage{amsmath,amssymb}
\usepackage{amsthm}
\usepackage{dsfont}
\usepackage{multirow}
\usepackage{multicol}
\usepackage{graphicx}
\usepackage{listings}
\usepackage{microtype}
\usepackage{graphicx}
\usepackage{wrapfig}
\usepackage{booktabs}
\usepackage{algorithm}
\usepackage[noend]{algpseudocode}
\usepackage{caption}
\usepackage{subcaption}
\usepackage{hyperref}
\usepackage{cleveref}
\usepackage{xcolor,colortbl}
\usepackage{xspace}
\usepackage{comment}
\usepackage{bm}
\usepackage{capt-of}    
\usepackage{tabularx} 
\usepackage{wrapfig}%
\usepackage{xr}

\newcommand{\mean}[2]{\mathbb{E}_{#2} \! \left[ #1 \right]}
\newcommand{\vect}[1]{\boldsymbol{#1}}
\newcommand\norm[2]{{\left\lVert#1\right\rVert}_{#2}}

\makeatletter
\newcommand{\printfnsymbol}[1]{%
  \textsuperscript{\@fnsymbol{#1}}%
}

\title{In search of dispersed memories: Generative diffusion models are associative memory networks}

\author{%
  Luca Ambrogioni \\
  Radboud University, Donders Institute for Brain, Cognition and Behaviour\\
  \texttt{luca.ambrogioni@donders.ru.nl} \\
}

\begin{document}

\maketitle

\begin{abstract}
Uncovering the mechanisms behind long-term memory is one of the most fascinating open problems in neuroscience and artificial intelligence. Artificial associative memory networks have been used to formalize important aspects of biological memory. Generative diffusion models are a type of generative machine learning techniques that have shown great performance in many tasks. Like associative memory systems, these networks define a dynamical system that converges to a set of target states. In this work we show that generative diffusion models can be interpreted as energy-based models and that, when trained on discrete patterns, their energy function is (asymptotically) identical to that of modern Hopfield networks. This equivalence allows us to interpret the supervised training of diffusion models as a synaptic learning process that encodes the associative dynamics of a modern Hopfield network in the weight structure of a deep neural  network. Leveraging this connection, we formulate a generalized framework for understanding the formation of long-term memory, where creative generation and memory recall can be seen as parts of a unified continuum. 
\end{abstract}

\section{Introduction}
Memory is a mysterious thing. Human beings can form lifelong memories from fleeting events and effortlessly recall them decades later as vivid multi-sensory experiences. On the other hand, in spite of their impressive capabilities, deep learning systems tend to require extensive training sessions to encode new information, which severely limit their adaptivity and consequently their capacity to develop general intelligence. Nevertheless, the field of machine learning has a long history of research on biologically plausible memory \citep{hopfield1982neural, michel1990associative, graves2016hybrid, lopez2017gradient, guo2020improved, kadam2020review}. Arising from the pioneering work of John Hopfield, associative memory networks have been proposed as computational models of biological memory \citep{hopfield1982neural, abu1985information}. These networks encode memories as stable fixed-points in an "energy landscape" defined on the space of neural activations. Interestingly, this energy function can be encoded into a pattern of synaptic connections trained using biologically plausible synaptic learning rules \citep{sejnowski1989hebb, michel1990associative}. In recent years, associative memory networks have been generalized in order to greatly scale their encoding capacity \citep{krotov2016dense, demircigil2017model}. However, these \emph{modern Hopfield networks} have a much more tenuous connection with known forms of synaptic learning since their energy function cannot be straightforwardly captured by learned pairwise synaptic couplings \citep{krotov2020large}. While biologically plausible implementations of modern Hopfield networks have been proposed, they do not provide strong insights on how memories are encoded in synaptic patterns since they require hard storage of the memorized pattern \citep{krotov2020large}. 

While human-like long-term memory is still outside of the capabilities of artificial intelligence systems, great progress has been made in approximating several forms of human creativity. Modern generative models are nowadays capable of generating beautiful visual art and insightful written text. In fact, these generative models have attracted great attention in neuroscience since they can provide an internal model of the world \citep{friston2006free, friston2010free, ororbia2022neural} and form the mechanistic basis for imagination and top-down predictive perception \citep{jones2020prediction}. Interestingly, modern research in psychology and neuroscience suggests that there is no sharp distinction between memory and spontaneous imagination \citep{schacter2007remembering}. In particular, the theory of reconstructive memory suggests that most sensory aspects of our memories are not encoded but are instead reconstructed from contextual information \citep{kolodner1983reconstructive, hemmer2009bayesian, schacter2007remembering, nash2015persuadability}. Memory-related brain areas such as the hippocampus have been shown to respond to imagination, future prediction and counterfactual reasoning tasks \citep{schacter2007remembering}. Moreover, hippocampal replays can be seen as a form of spontaneous generation, which is considered to be vital for memory consolidation and planning \citep{buhry2011reactivation}. Given these deep connections, it is natural to expect that generative models and associative memory models are two faces of the same coin. 

In this paper, we will show that this is indeed the case for generative diffusion models, a relatively new class of generative models that have achieved state-of-the-art performance in most computer vision and audio generation tasks \citep{sohl2015deep, song2020denoising, ho2020denoising}. Specifically, we show that the asymptotic low-time energy landscape of a large class of generative diffusion models trained on discrete patterns is identical to the asymptotic low-temperature landscape of modern Hopfield networks. Furthermore, in our experiments, we show that this equivalence holds almost exactly in the standard numerical implementations. 

Using these results, we offer a new theoretical conceptualization of associative memory that can incorporate semantic, episodic and reconstructive memory as the result of the action of the same synaptic learning rule. 

\section{Preliminaries}
In this section we will review the basic theory of associative memory and diffusion modeling. To keep the text readable to a large audience, we will focus on the intuitive aspects and keep the level of mathematical formalism at a minimum. For example, we will write stochastic differential equations in terms of (infinitesimal) update equations. For a more formal treatment of these topics, we recommend the reader to refer to SDE texts such as to \citep{kloeden1992stochastic}.

\subsection{Associative memory networks}
Hopfield networks have been developed to formalize the concept of associative memory in a simplified artificial neural system \citep{hopfield1982neural}. We will denote the activity of the $D$ memory units with a vector $\vect{x}(t) = (x_1(t), \dots,x_D(t))$. In the original paper, these neural activities where assumed to be binary variables ($x_j(t) \in \{-1,1\}$), respectively denoting states of rest and states of firing. The dynamic of a Hopfield network is regulated by the update equation:
\begin{equation}
    x_j(t+\text{d}t) = \left[\text{sign}\left(W \vect{x}\right) \right]_j~,
\end{equation}
where $W$ is a real-valued symmetric matrix of synaptic couplings (weights) with null diagonal ($W_{jj} = 0,~ \forall j$). The matrix $W$ encodes pairwise associations between the different components of the pattern vectors. It can be show that this update rule decreases monotonically the following energy function \citep{hopfield1982neural}:
\begin{equation}
    u_\text{H}(\vect{x}) = \vect{x}^T W \vect{x}~,
\end{equation}
and that it will therefore converge to one of its local minima. These minima can then be used to encode memories, which can be retrieved through the Hopfield dynamics when initialized in an incomplete or perturbed version of the memory state. The simplest way to encode memories in the coupling matrix is to use the Hebb's rule of association, which in its simplest form is 
\begin{equation}
    W_{j,k} = \sum_{n=1}^N y_{n,j} y_{n,k}
\end{equation}
where the set of vectors $\{\vect{y}_1, \dots, \vect{y}_n, \dots \vect{y}_N\}$ represents $N$ "experienced" patterns of neural activity. A pattern is considered to be successfully stored if it is a stable fixed-point of the discrete dynamics. If the patterns are random, it can be proven that the storage capacity of Hopfield nets scales as ${D}/{4 \log_2 D}$ \citep{abu1985information}. 

The storage capacity of Hopfield-like networks can be increased by including non-linear mappings $F(\cdot)$ in its energy function \citep{krotov2023new, krotov2016dense, demircigil2017model}. The general form for the energy of a (discrete) modern Hopfield network can be written as 
\begin{equation} \label{eq: modern Hopfield energy}
    u_{F}(\vect{x}) = h\left( \sum_{n=1}^N F\left(\vect{x}^T \vect{y}_n \right) \right)~,
\end{equation}
where $h(\cdot)$ is an arbitrary differentiable and strictly monotonic function, which does not affect the location and stability of the local minima. This expression reduces to the standard Hopfield energy for $F(x) = x^2$ and $h(x) = x$. However, it is possible to achieve much higher theoretical capacity by using more complex functions. For example, a modern Hopfield network with $F(x) = e^x$ can store up to $2^{D/2}$ binary patterns \citep{demircigil2017model}.  These associative networks have been recently generalized to have continuous dynamics. If the activation vector is continuous, the energy function needs to include a regularization term to enforce stability. For example, \citep{ramsauer2020hopfield} proposed the use of an energy function of the form
\begin{equation} \label{eq: continuous Hopfield energy}
    u_{\text{MH}}(\vect{x}, \beta) = - \beta^{-1} \log \left(\sum_{n=1}^N e^{\beta \vect{x}^T \vect{y}_n} \right) + \norm{\vect{x}}{2}^2/2
\end{equation}
where $\beta$ is a positive-valued parameter. We omitted the terms that are additive constant in $\vect{x}$ since they do not change the fixed-points and the resulting dynamics. \citep{krotov2020large} showed that this dynamics can be expressed in terms of biologically plausible binary association between latent neurons, in a way that is similar to the architecture of a restricted Boltzmann machine \citep{fischer2012introduction}. Unfortunately, in these models, the numerical values of the patterns directly determine the synaptic weights between latent and observable neurons. This implies that the patterns need to be stored in memory instead of being converted into distributed synaptic patterns. In this sense, modern Hopfield networks offer a model of memory recall but do not provide insight into learning and memory storage in the brain. 

\subsection{Generative diffusion models} \label{sec: diffusion models}
Consider a target  distribution $\phi(\vect{y})$ that is only available through a training dataset $\mathcal{D} = \{\vect{y}_1, \dots, \vect{y}_N \}$ of independently sampled data-points. Our goal is to learn the structure of the training set in order to generate new samples from $\phi(\vect{y})$. To this aim, we first define a noise-injective process that turns the training samples into random noise states. We will then 'invert' this process to turn random noise into new samples. To be consistent with the notation used in the continuous Hopfield model, we deviate from the diffusion modeling literature by writing this process in reversed time, with the noiseless data corresponding to a final time $T$. In this notation, the diffusive dynamics can be determined by the following backward recursive equation:
\begin{equation}
\label{eq: noise-injection}
    \vect{x}(t - \text{d}t) = \vect{x}(t) + \sigma \sqrt{\text{d} t} \vect{\delta}(t)~,
\end{equation}
where $\sigma$ determines the standard deviation of the noise injected at time $t$ and $\vect{\delta}(t)$ follows a standard normal distribution. This is known as the variance exploding equation in the generative diffusion modeling literature \citep{song2021scorebased}, which corresponds to a mathematical Brownian motion. We can sample from this process by sampling an initial state $\vect{y}$ from the dataset and using it as the final state in the recursion defined by Eq.~\ref{eq: noise-injection}. Note that any other Ito diffusion process defined by a stochastic differential equation can equivalently be used to define a generative diffusion model \citep{song2021scorebased}, we will cover the general case in section \ref{sec: general case}.  It can be shown that, if Eq.~\ref{eq: noise-injection} is initialized with the target distribution $\phi(\vect{y})$, the 'inverse' equation is given by 
\begin{equation}\label{eq: generative dynamics}
    \vect{x}(t + \text{d}t) = \vect{x}(t) + \sigma^2 \nabla_{\vect{x}} \log{p_{t}(\vect{x}(t))} \text{d} t + \sigma \sqrt{\text{d} t} \vect{\delta}(t)~. 
\end{equation}
where $p_t(\vect{x})$ is the marginal distribution of the noise-injection process at time  $t$. In the case of the variance-exploding process, the marginal can be computed analytically and it can be expressed as
\begin{equation} 
    p_{t}(\vect{x}) = \mean{\frac{1}{\sqrt{2 \pi (T - t) \sigma^2 }} e^{-\frac{\norm{\vect{x} - \vect{y}}{2}^2}{2 (T - t) \sigma^2}}}{\vect{y} \sim \phi(\vect{y})}~.
\end{equation}
This formula involves an average over the distribution of the data, which is usually not available in a generative modeling task. However, we can approximate the drift term of the dynamics (i.e. the so called score $\nabla_{\vect{x}} \log{p_{t}(\vect{x}})$) using a parameterized deep network $\vect{s}(\vect{x}(t), t; W)$ trained on the denoising loss \citep{song2020denoising}:
\begin{equation}\label{eq: diffusion learning loss} 
    \mathcal{L}(W) = \frac{1}{2} \mean{\mean{\norm{\vect{\delta}(t) - \vect{s}(\vect{x}(t), t; W)}{2}^2}{\vect{x}(t) \mid \vect{y}}}{\vect{y}\sim \mathcal{D} ), t}
\end{equation}
where $\vect{\delta}(t) = \vect{x}(t) -  \vect{y}$ is the total noise added to the pattern $\vect{y}$ up to time $t$. The score can then be recovered from the network using the formula
\begin{equation}
    \nabla_{\vect{x}} \log{p_{t}(\vect{x})} \approx - \sigma^{-1} \vect{s}(\vect{x}(t), t; W)
\end{equation}
which becomes exact when the dataset is infinitely large and the loss is minimized globally. 


\section{The equivalence between diffusion models and modern Hopfield networks}
\begin{figure}
\centering
\includegraphics[width=13cm]{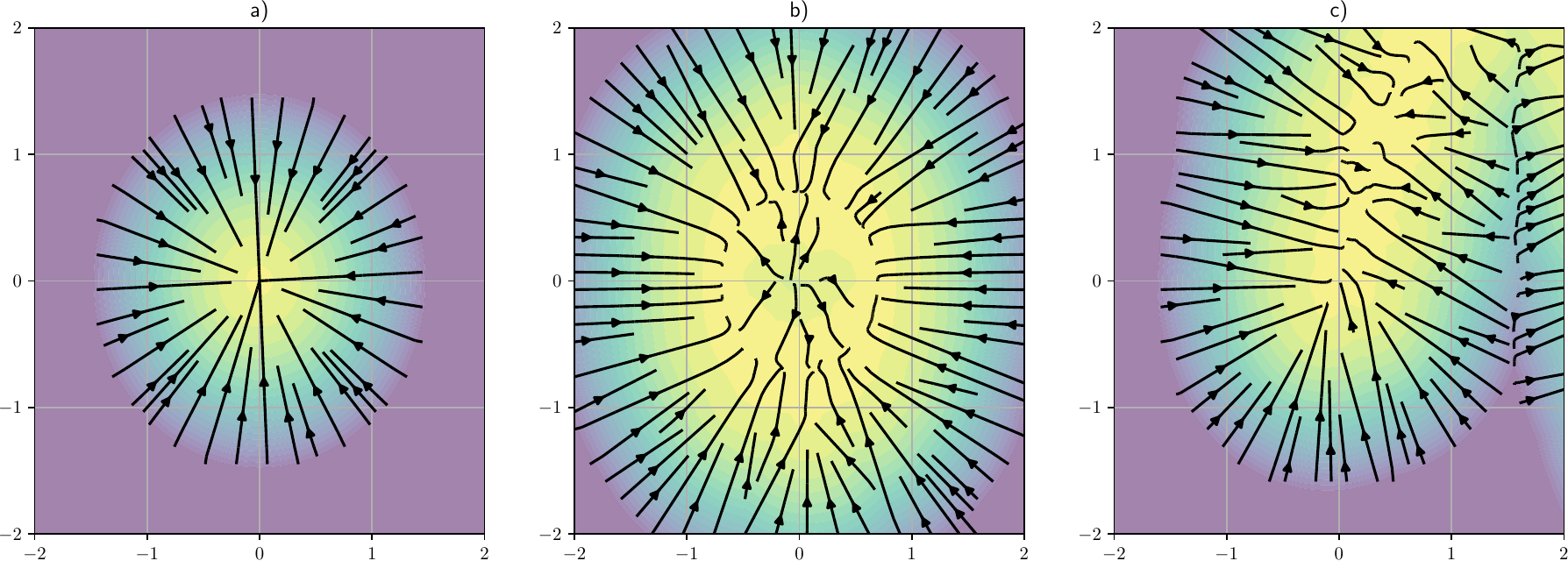}
\caption{Visualization of different kinds of energy landscape and gradient vector fields corresponding to different forms of memory (in a two-dimensional space): a) Classical point-like memory; b) Extended localized memory; c) Non-localized (semantic) memory structure. }
\label{fig: landscape}
\end{figure}
We can now show that, when used for storing discrete patterns, the dynamics of generative diffusion models minimizes the energy function of continuous modern Hopfield networks. We will start from the simpler case of variance-exploding diffusion models as their analysis provides all the main insights and results. We will then generalize the result to arbitrary diffusion models defined by a large class of SDEs. 

\subsection{Variance-exploding models} \label{sec: exploding case}
The first step is to formulate the deterministic dynamics of the model as the negative gradient of an energy function:
$
    \vect{x}(t + \text{d}t) = \vect{x}(t) - \nabla_{\vect{x}} u(\vect{x}, t) \text{d} t + \sigma \sqrt{\text{d} t} \vect{\delta}(t)~.
$
As noted in \citep{raya2023spontaneous}, the energy function is
\begin{equation} \label{eq: energy}
    u_{\text{DM}}(\vect{x}, t) = - \sigma^2 \log{p_{t}(\vect{x})} = - \sigma^2 \log{\mean{e^{-\frac{\norm{\vect{x} - \vect{y}}{2}^2}{2 (T - t) \sigma^2}}}{\vect{y} \sim \phi(\vect{y})}} + c~,
\end{equation}  
where $c$ does not depend on $\vect{x}$ and can therefore be omitted without affecting the dynamics. In order to establish a link between the diffusion model and Hopfield networks, we can now assume that the data source is a finite collection of $N$ patterns that we wish to store as memories. This lead to the energy
\begin{equation} \label{eq: diffusion energy}
    u_{\text{DM}}(\vect{x}, t) = - \sigma^2 \log{\left(\frac{1}{N} \sum_{n = 1}^N {e^{-\frac{\norm{\vect{x} - \vect{y}_n}{2}^2}{2 (T - t) \sigma^2}}}\right)}~.
\end{equation}
If we now assume that the patterns are normalized ($\norm{\vect{y}}{2}^2 = 1$), by expanding the square in the exponent and omitting constant additive terms, we obtain
\begin{equation}
    u_{\text{DM}}(\vect{x}, t)/\sigma^2 = - \log{\left(\sum_{n = 1}^N {e^{\frac{\vect{x}^T \vect{y}_n}{2 (T - t) \sigma^2}}}\right)} + \frac{\norm{\vect{x}}{2}^2}{(T - t)^2}~.
\end{equation}\
Finally, if we define $\beta(t)^{-1} = (T - t) \sigma^2$ and we multiply both sides by $\beta(t)^{-1}$, we obtain
\begin{equation} \label{eq: hopfield-like energy}
    \beta^{-1}(t) u_{\text{DM}}(\vect{x}, t)/\sigma^2 = - \beta(t)^{-1} \log{\left(\sum_{n = 1}^N {e^{\beta(t) \vect{x}^T \vect{y}_n}}\right)} + \frac{\norm{\vect{x}}{2}^2}{2} = u_{\text{MH}}(\vect{x}, \beta(t))~, 
\end{equation}
which for a fixed $t$ is identical to the continuous Hopfield network energy in Eq.~\ref{eq: continuous Hopfield energy}, and it therefore has the same fixed-point structure at the limit $\beta \rightarrow \infty$. Note that the scaling factor $\beta^{-1}(t)/\sigma^2$ does not change the fixed-points and their stability as it is a positive constant of $\vect{x}$. While we derived this result assumes normalized patterns, this is not actually necessary since, for $\beta(t) \rightarrow \infty$, we have that
$
    \beta^{-1}(t) u_{\text{DM}}(\vect{x}, t)/\sigma^2 \sim - \frac{1}{2} \left( \vect{x}^T \vect{y}^* + \norm{\vect{y}^*}{2}^2 + \norm{\vect{x}}{2}^2 \right)
$
where $\vect{y}^*$ is the pattern that maximizes the quadratic form $\vect{x}^T \vect{y} + \norm{\vect{y}}{2}^2$. As shown by this expression, at this limit the norm $\vect{y}^*$ only adds an irrelevant constant shift in the energy.  

The main differences between the two approaches are 1) in diffusion models, $\beta(t)$ tend to this divergent limit as part of the denoising dynamics, while the denoising iterations of modern Hopfield networks keep $\beta$ fixed.; and 2) in Hopfield network the energy function is minimized deterministically while in diffusion models these is an additional stochastic term. However, these two differences 'cancel each other out' since the divergence of $\beta(t)$ leads to the suppression of the stochastic fluctuations and to exact convergence on the same patterns that minimize the modern Hopfield energy for $\beta \rightarrow 0$. As shown in our experiments, there is no meaningful difference as far as $\beta$ is large enough.

\subsection{The general case} \label{sec: general case}
We can now prove the equivalence in a much more general case where the noise injection dynamics follows the equation
\begin{equation}
    \vect{x}(t - \text{d}t) = \vect{x}(t) + \nabla v(\vect{x}) \text{d}{t} + \sigma(t) \sqrt{\text{d} t} \vect{\delta}(t)~,
\end{equation}
where $v(\vect{x})$ is a differentiable scalar potential function and $\sigma(t)$ is a continuous function of time. This form covers most of the equations used in the diffusion modeling literature but it excludes non-conservative dynamics and state-dependent noise models. The generative dynamics corresponding to this more general noise-injection model is given by the equation
\begin{equation}\label{eq: general generative dynamics}
    \vect{x}(t + \text{d}t) = \vect{x}(t) + \left(\sigma(t)^2 \nabla_{\vect{x}} \log{p_{t}(\vect{x}(t))} - \nabla v(\vect{x}) \right) \text{d} t + \sigma(t) \sqrt{\text{d} t} \vect{\delta}(t)~. 
\end{equation}
where the conditional marginal distributions of the stochastic differential equation is given by the expression
\begin{equation}\label{eq: conservative equation}
    p_{t}(\vect{x}) = \mean{k(\vect{x}(t), t; \vect{y}, T)}{\vect{y} \sim \phi(\vect{y})}~.
\end{equation}
In this formula, the solution kernel $k(\vect{x}'', t''; \vect{x}', t')$ gives the conditional probability density of a state $\vect{x}'$ at $t'$ to be moved to $\vect{x}''$ at time $t''$ under the noise-injecting dynamics. Unfortunately, this solution kernel cannot be expressed analytically in the general case. We define the function $\psi(\vect{x}, t; ~\vect{y}) = \log{k(\vect{x}, t; \vect{y}, T)}$ as the logarithm of the solution kernel, this allows us to express the energy function of the model in the following form
\begin{equation}
    u_{v\text{DM}}(\vect{x}, t) = - \sigma^2(t) \log{\left( \sum_{n = 1}^N e^{\psi(\vect{x}, t; ~\vect{y}_n)} \right)} + v(\vect{x})~
\end{equation}
As we showed in the previous section, the equivalence with Hopfield model can be shown at the limit $t \rightarrow T$, which corresponds to $\beta \rightarrow \infty$. At this limit, the diffusion dynamics of Eq.~\ref{eq: conservative equation} around a pattern $\vect{y}^*$ is well-approximated by the linearized equation
\begin{equation}
    \vect{x}(t - \text{d}t) - \vect{x}(t) = \nabla v(\vect{y}*) \text{d}{t} + H(\vect{y}*) (\vect{x} - \vect{y}^*) \text{d}{t} + \sigma(t) \sqrt{\text{d} t} \vect{\delta}(t)~,
\end{equation}
with $H(\vect{y}*)$ denoting the Hessian matrix of the potential at $\vect{y}*$, with $H_{ij}(\vect{y}*) = \frac{\partial^2 u(\vect{y}^*)}{\partial x_j \partial x_j}$. Since the equation is linear, its solution kernel is given by a Gaussian density whose mean vector and covariance matrix depend on the potential $v(\vect{x})$ and on $\sigma(t)$. For $t \rightarrow T$, the drift term is negligible compared to the diffusion term as the former scales linearly while the latter scales as a squared root. For this reason, the kernel simplifies further into the following asymptotic expression
\begin{equation}
    \psi(\vect{x},t; \vect{y}) \sim -\frac{1}{2 \sigma^2(T) (T - t)} \norm{\vect{x} - \vect{y}}{2}^2 + c~.
\end{equation}
where $c$ does not depend on $\vect{x}$. Up to multiplicative and additive  constants, this leads to the asymptitic energy function
\begin{equation}
    u_{v\text{DM}}(\vect{x}, t) \sim - \log{\left( \sum_{n = 1}^N {e^{-\frac{\norm{\vect{x} - \vect{y}_n}{2}^2}{2 (T - t) \sigma(T)^2}}}\right)} ~,
\end{equation}
which is identical to the energy function in Eq.~\ref{eq: diffusion energy}. Note that we neglected the "regularization term" $v(\vect{x})$ since it does not diverge for $t \rightarrow T$ and it is therefore negligible at this limit. From this result, we can conclude that the fixed-point structure of the diffusion model at $t \rightarrow T$ does not depend on the specific form of the SDE and it is therefore fully characterized by the Brownian motion model analyzed in Sec.\ref{sec: exploding case}.

\section{Encoding memories by denoising neural state}
So far, we considered exact diffusion models by analyzing the theoretically optimal score function. However, in real-world applications it is often not possible to compute the score analytically. Instead, as outlined in Sec.~\ref{sec: diffusion models}, the score function is approximated using a denoising deep neural network $\vect{s}(\vect{x}(t), t; W)$ parameterized by a large array $W$ of synaptic weights. Therefore, in real applications the memories are ultimately encoded in the pattern of synaptic weights updated by SGD. Specifically, for a single data-point, the loss function in Eq.~\ref{eq: diffusion learning loss} results to a SGD weight updates of the form
\begin{equation} \label{eq: synaptic update}
    W_{t+1} = W_t - \eta \frac{\partial \vect{s}}{\partial W} \bigg( \vect{\delta}(t) - \vect{s}(\vect{x}, t; W) \bigg) ~.
\end{equation}
where $\vect{\delta}(t)$ can be interpreted as the residual between a past low-noise state $\vect{y}$ and a current noisy-state $\vect{x}$. In the brain, the residual can be obtained through short-term sensory memory, which can buffer recent states and then compare them with the current activity. While the activity of the brain is ongoing and cannot be neatly separated into forward and reversed dynamics, there is tantalizing evidence that the time-evolution of cortical networks oscillate between feed-forward and feed-back phases \citep{hasselmo1996suppression}, which corresponds to states with different signal-to-noise ratios \citep{hasselmo1997noradrenergic}. This phenomenon might offer a mechanistic implementation of generative diffusion training in the brain, possibly in relation with the hippocampus theta cycle, which is known to play a crucial role in memory consolidation \citep{hasselmo2000septal}. Interestingly, recent studies show that the backpropagation term ${\partial s}/{\partial W}$ can also be obtained through noise injection \citep{dalm2023effective}, which suggests that noise-injecting processes may play a major role in learning. 

\section{Beyond classical associative memories} 
In the previous section, we showed that the asymptotic energy landscape of generative diffusion models trained on a finite set of discrete patterns matches the energy function of modern Hopfield networks for $\beta \rightarrow \infty$. This implies that, in this regime, the two models have the same fixed-point structure and, consequently, that they have the same memory capacity. Nevertheless, generative diffusion models are more general than associative memory networks in two respects: 1) they provide a framework for probabilistic memory recall and 2) they can be used to encode non-discrete memory structures of arbitrary dimensionality. In this sense, generative diffusion model theory can be seen as a wide generalization of the classical theory of associative memory. 

\subsection{Probabilistic recall}
At its core, the classical theory of associative memory is centered around the idea of deterministic energy minimization (however, see \cite{hancock1993bayesian} for related probabilistic methods). On the other hand, the theory of generative diffusion modeling is based on partially random processes. This is not a coincidence since these models have been devised for the purpose of generation and statistical variability is central to the generative task. However, probability theory is also central to the problem of memory retrieval, since in general there are many possible patterns $\vect{y}$ that could have generated a given corrupted/incomplete state $\vect{x}$. This is particularly central when the level of corruption is high (i.e. for large values of $(T - t)$), since in this regime it is often impossible to distinguish between several possible patterns and the best hope is to cover their (posterior) probability distribution. When initialized in one of these partially-corrupted states, the stochastic generative dynamics of diffusion models can be interpreted in this fashion, with each denoising trajectory eventually reaching one of the possible fixed-points that are compatible with the initial state. Probabilistic recall can be used to encode Bayesian uncertainty, with several possible memories being simultaneously co-activated by parallel denoising processes. This form of encoding can be used to solve information gathering, exploration and other meta-learning problems \citep{gupta2018meta} and has been suggested to be central to the functioning of the mammalian hippocampus \citep{ambrogioni2023rethinking}. 

\subsection{Higher-dimensional memory structures}
In the classical associative memory literature, a memory is encoded as a single (fixed-)point in a $m$-dimensional space of possible neural activities. In modern mathematics, a set of points is often interpreted as a $0$-dimensional space as there are no local degrees of freedom, meaning that it is impossible to locally perturb a point while remaining inside its 'space'. The energy landscape and associated vector field of this form of memories is visualized in Fig.~\ref{fig: landscape} a. This presents a very strict definition of memory that is incompatible with its cognitive reality in most biological and artificial forms of intelligence. In fact, even the sharpest human memory only encodes a minority of the details of the original neural state, with many of these details being reconstructed during recall based on contextual information \citep{nash2015persuadability}. This suggests that a more realistic description of memory should include "internal" degrees of freedom that allow for partial encoding. In mathematical terms, this can be done by defining a memory as a connected $d$-dimensional sub-space embedded in the larger $m$-dimensional space of possible neural activities. A diffusion model trained on $N$ of these "extended memories'', each represented by a sub-space $\mathcal{S}_n$ has the energy function
\begin{equation} \label{eq: extended memories}
    \log{\left( \sum_{n = 1}^N \int_{\mathcal{S}_n} {e^{ \beta(t) \vect{x}^T \vect{y}_n}} \text{d} \vect{y}_n \right)}~,
\end{equation}
where the integral is taken over the whole sub-space $\mathcal{S}_n$, whose points correspond equally valid interpretations of the same memory. An example of an energy landscape for this form of extended localized memory is given in Fig.~\ref{fig: landscape} b. In general, each sub-space $\mathcal{S}_n$ can potentially have a different dimensionality or even be a more complex geometrical structure with fractional (fractal) or variable dimensionality.

\section{A theoretical framework for memory in biological and artificial intelligent systems}
In this section we will leverage the connection we established between associative memory and diffusion models in order to outline a theoretical framework for biologically plausible memory. Our goal is both to provide conceptual tools to theoretical and experimental neuroscientists and to promote developments in naturalistic machine intelligence systems.

\subsection{Semantic, episodic and reconstructive memory}
In psychology and neuroscience, the semantic (or structural) memory system is thought to learn the general structure of the sensory input, discarding the idiosyncratic details of individual events. This can be modeled with an energy function where the sum over patterns is replaced by an integral over a continuous density $\phi(\vect{y})$ :
\begin{equation} \label{eq: semantic memories}
    \sigma^2 \log{\int e^{\beta(t) \vect{x}^T \vect{y}} \phi(\vect{y}) \text{d} \vect{y} }~.
\end{equation}
In practice, the distribution $\phi(\vect{y})$ is often defined on a manifold or some other lower dimensional structure, leading to a dynamics like what is visualized in Fig.\ref{fig: landscape} c. This is exactly the kind of behavior we expect from a generative model, initial perturbed states are gradually pushed towards a point on the manifold of possible patterns. In a sense, it can also be seen as a probabilistic generalization of the high dimensional memory structures described in Eq.~\ref{eq: extended memories}, where the sub-space $\mathcal{S}_n$ is no longer localized in a small sub-region of the space of neural states. In cognitive terms, roughly speaking we can say that a memory is "episodic" is its sub-space is localized while that a memory is semantic or structural if it is part of a large non-localized sub-space. 

However, modern memory research suggests that real-life episodic memories are thought to be largely reconstructive, meaning that most of the sensory details are re-created during recall based on contextual information \citep{roediger1995creating, hemmer2009bayesian}. For example, the memory of a car crash may evoke the memory of broken glass, although the windshield was not actually broken during the real event. This suggests that human episodic memory has a stored lower dimensional "representational core" that does not fully constrains the dynamics of the system. This can be formalized using a mixture of discrete (or localized) and continuous distributions:
\begin{equation}\label{eq: mixture energy}
    \log{\left(\sum_{n}^N e^{\beta(t) \vect{f}_n(\vect{x})^T \vect{\xi}_n} + \int e^{-\beta(t) \vect{x}^T \vect{y}} \phi(\vect{y}) \text{d} \vect{y} \right)}~,
\end{equation}
where the function $\vect{f}_n: \mathds{R}^D \rightarrow \mathds{R}^W$ with $W < D$ is a lower dimensional encoding of the state and $\vect{\xi}^n \in \mathds{R}^W$ is a stored lower-dimensional pattern that encode the "core" of the memory. Since $W < D$, this energy will only constrain $W$ degrees of freedom to converge to the pattern while the other degrees of freedom are left free to evolve under the dynamics determined by the corpus of semantic memory. This leads to a form of memory recall where some features are recovered faithfully while others are reconstructed based on learned contextual associations. 

\subsection{Consolidation and replays}
All the different forms of memory outlined in the previous section can be learned through the same synaptic update rule given in Eq.~\ref{eq: synaptic update}. In this sense, our model provides a possible unification for most of the known-form of long-term memory in humans and other animals. During training, the only difference between learning a density $\phi(\vect{y})$ or a discrete set of patterns is that in the former case each pattern $\vect{y}$ is (almost surely) sampled only once while in the latter case each pattern is re-sampled with finite probability throughout training. In generative machine learning, re-sampling is often performed in practice to maximize data efficiency, although it can lead to overfitting since the generative model could memorize the patterns themselves instead of extrapolating the underlying density $\phi(\vect{y})$. 

In biological systems, each data-point can only be experienced once and this would likely not result in the formation of localized episodic memories, since their encoding requires some form of re-sampling. However, it is well-known that the formation of episodic memories in humans requires an extensive phase of consolidation that depends on the activity of the hippocampus. Under our theoretical framework, a possible explanation is that new simulated experiences are generated in the hippocampal network and are then used for synaptic training. In fact, it is well-known that these forms of replays can be observed in the hippocampus both during sleep and wakefulness and that their disruption compromises memory consolidation \citep{olafsdottir2018role}. Training on these self-generated samples can result in a positive feedback, with each reply increasing the probability of the same event being re-sampled in the future. Through this process, a single event can be 'bootstrapped' into a self-reinforcing process of activity that can eventually form a localized episodic memory. Furthermore, this would result in the formation of a low-dimensional episodic "core", as described in the previous sub-section, if only a subset of features are re-sampled faithfully, with the other being generated from structural memory. 

\begin{figure}
\centering
\includegraphics[width=12cm]{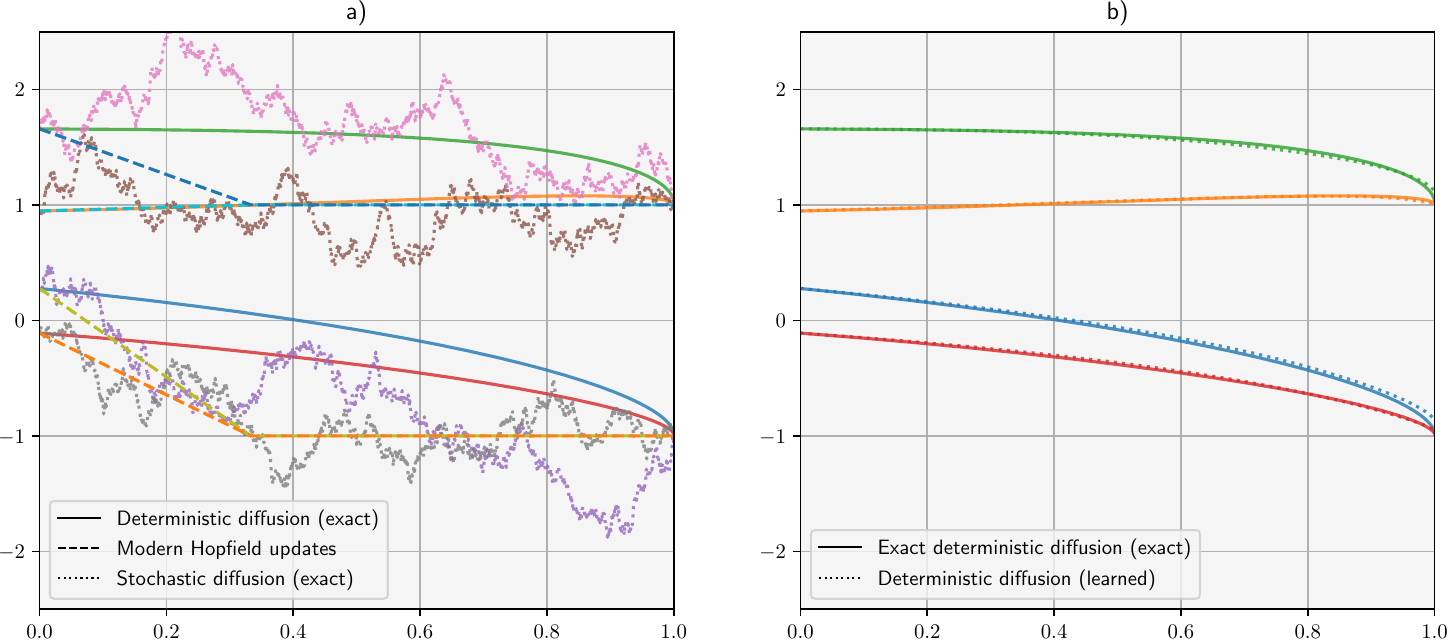}
\caption{Qualitative analysis of the (marginal) denoising trajectories of a binary associative memory problem with $4$ patterns in a $5$ dimensional space. a) Comparison between denoising trajectories of diffusion models and modern Hopfield updates. The diffusion curves are integrated using the Euler method with $2000$ steps. The trajectories are overlaid to four modern Hopfield updates. b) Comparison between exact and learned deterministic denoising trajectories. }
\label{fig: denoising comparison}
\end{figure}

\section{Experiments}
Fig.\ref{fig: denoising comparison} visualizes the qualitative behavior of both learned and exact score models compared with modern Hopfield iterations. We consider a $5$-dimensional associative memory problem with $4$ randomly sampled binary patterns. For the forward diffusion dynamics, we used a variance-preserving model as they are numerically more stable and more widely used \citep{song2020denoising}. We considered both exact score models and trained diffusion models parameterized by three layer perceptrons with $750$ hidden units in each hidden layer. We used the training approach described in \citep{song2020denoising} but with a simpler constant noise schedule ($\sigma$ = 1). Diffusion trajectories were integrated using a simple Euler approach for the deterministic ODE samples (see \citep{song2021scorebased}, while we used a Euler–Maruyama method for the SDE sampler. The modern Hopfield iterations were implemented as specified in \citep{ramsauer2020hopfield}, with $\beta = 5$ and $4$ updates.  As expected from our analysis, all algorithms approximately converge to the same target points. As expected from previous work, the modern Hopfield iteration converges after a single iteration while the trajectories of the diffusion trajectories converge smoothly to the target. Note however that, in this setting, diffusion denoising can also be performed in one step simply by re-scaling the score. 

As first quantitative analysis we evaluated the Pearson correlation coefficient between the output of modern Hopfield iteration and a) a diffusion model, b) a classical Hopfield network and c) the ground truth pattern. For a given dimensionality $d$, $n$ binary patterns $\vect{y}$ were randomly generated and subsequently corrupted with noise using the formula $\tilde{\vect{y}} = \theta \vect{y} + \sqrt{1 - \theta^2} \epsilon$, where $\epsilon$ is a standard Gaussian noise vector. We used a noise level of $\theta = 0.68$.  We kept the dimensionality equal to $10$ and we evaluated the correlation for $10$, $20$ and $30$ stored patterns. The simulation was repeated $100$ times in order to reliably compute the correlations. In order to avoid to have to re-train a neural model hundreds of times, for the diffusion models we used the exact score formula (see Supp.\ref{supp: experimental details}. For the modern Hopfield model, we used $150$ updates in order to maximize performance. The final output of all methods was binarized using the sign function. Table \ref{tab: table 1} shows the estimated Pearson correlations between methods. As expected from our analysis, the correlation between the modern Hopfield iterations and the diffusion model is extremely close to one.  

\begin{table}[h]  \centering
\begin{tabular}{lllll} \toprule
\hline 
 Denoising task & \# Patterns & Diffusion Model & Classic Hopfield  & True patterns \\ \hline
& 10                      & 0.995    & 0.732            & 0.893         \\
& 20                       & 0.991     & 0.704       & 0.822         \\
& 30                      & 0.991   & 0.715         & 0.81         \\ \hline

 Completion task &  &  &   &  \\ \hline
& 10                      & 0.996    & 0.741            & 0.897        \\
& 20                       & 0.991     & 0.707       & 0.838        \\
& 30                      & 0.989   & 0.700         & 0.795       \\ \hline
\end{tabular} \caption{Pearson correlation between output of modern Hopfield network and other models (plus ground truth pattern) in both denoising and completion experiments.}
\label{tab: table 1}
\end{table}

We also performed the same experiments in a completion task, where the patterns were partially zero-masked instead of being corrupted by white noise. The binary masks were sampled randomly from a Bernoulli distribution with $p = 0.5$. Again, the output of the exact diffusion models correlates almost perfectly with the output of the modern Hopfield iterations. 

\begin{figure}
\centering
\includegraphics[width=12cm]{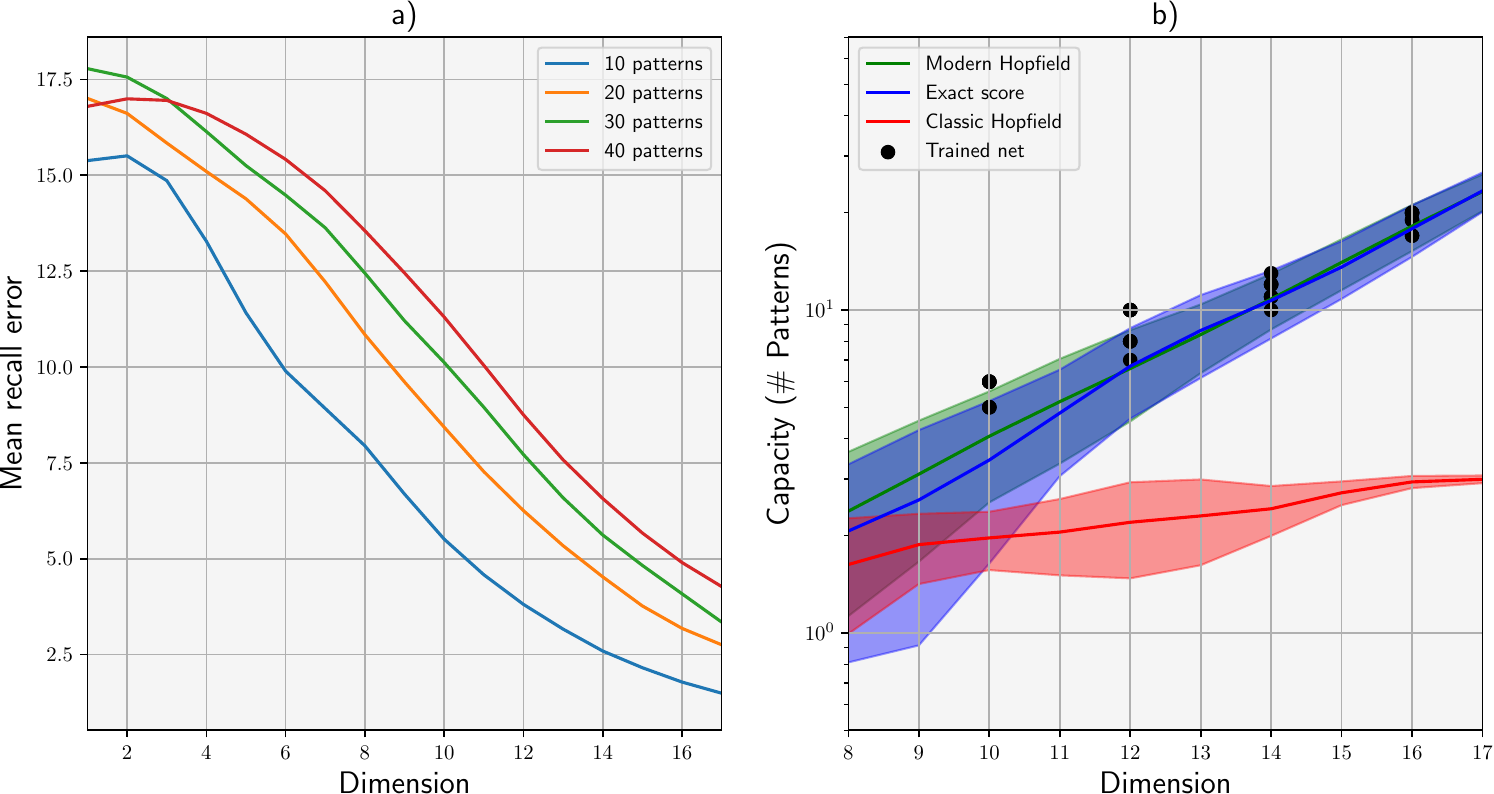}
\caption{1) Median error of exact diffusion model as function of the dimensionality. b) Capacity of diffusion models and Hopfield networks in log scale. The shaded area denotes the estimated $95\%$ intervals.}
\label{figure 2}
\end{figure}

Next, we estimated the error and capacity of the models. The corrupted patterns were fed to the algorithms and the results were compared with the original pattern using the Hamming error. The patterns were considered to be correctly recovered if the error was smaller than 3\%. Fig.~\ref{figure 2}a shows the error of an exact diffusion model for different numbers of patterns as function of the dimensionality. For a given noise level and threshold, the capacity was defined as the maximum number of patterns that can on average be recovered. Fig.~\ref{figure 2}b shows the estimated capacity of the exact diffusion model (blue), modern Hopfield network (green), classical Hopfield network (red) and trained diffusion model (black dots). The details of the experiments are given in Supp.~\ref{supp: experimental details}.

\section{Discussion}
In this paper, we demonstrated that a popular class of modern Hopfield networks with exponential non-linearities is mathematically equivalent to a large class of continuous diffusion models at the limit of $\beta \rightarrow \infty$. In our experiments, we showed that this equivalence holds almost exactly for a finite $\beta$ value and for the more stable variance-preserving models, both with exact and trained score. The equivalence depends on the fact that the diffusion models are trained on a finite number of discrete patterns, and in fact the diffusion models can generalize the modern Hopfield energy function to setting where both episodic memory and semantic memory (i.e. generative manifolds) are jointly encoded by the dynamics of the same network. From the point of view of theoretical neuroscience, this may be used to model the different forms of long-term memory as a result of the same learning mechanism, in a single distributed neural system. 

While generative diffusion models offer an attractive paradigm for modeling memory, imagery and even perception in the brain, more work needs to be done in order for its components to be implemented in a biologically plausible way. In particular, it is unlikely that biological neural networks implement pure noise-injection dynamics. However, the mathematics of generative diffusion models can be written in term of any stochastic differential equation \citep{song2021scorebased}, which can be used to implement more plausible transformation such as, for example, the feedforward perceptual feature as implemented in the hierarchy of sensory cortical areas \citep{gucclu2015deep}. More fundamentally, it is unclear how the reverse and forward dynamics can be simultaneously implemented in the brain networks. However, there is tantalizing evidence that this could correspond to the theta rhythm driving periodic cholinergic modulations \citep{hasselmo2000septal}.

The denoising loss given by Eq.~\ref{eq: diffusion learning loss} is somewhat similar to the self-prediction errors used in predictive coding models \citep{rao1999predictive, spratling2017review, millidge2021predictive}. This opens the door for potentially fruitful connections with existing predictive models of memory \citep{stachenfeld2017hippocampus, barron2020prediction}. Particularly interesting is the recent use of generative predictive coding models for associative memory storage \citep{salvatori2021associative}, which is also based on a generative model and has some similarities with the approach discussed here.

The main biological issue with episodic memory as conceptualized in this paper is that it seems to require the re-sampling of the same events, while in the real world each event can only happen once. This can be potentially implemented with some form of bootstrap re-sampling, which might correspond to the replays observed in the hippocampus \citep{buhry2011reactivation}. 

\bibliographystyle{apalike}

\bibliography{references.bib}

\begin{thebibliography}{}

\bibitem[Abu-Mostafa and Jacques, 1985]{abu1985information}
Abu-Mostafa, Y. and Jacques, J.~S. (1985).
\newblock Information capacity of the hopfield model.
\newblock {\em IEEE Transactions on Information Theory}, 31(4):461--464.

\bibitem[Ambrogioni and {\'O}lafsd{\'o}ttir, 2023]{ambrogioni2023rethinking}
Ambrogioni, L. and {\'O}lafsd{\'o}ttir, H.~F. (2023).
\newblock Rethinking the hippocampal cognitive map as a meta-learning computational module.
\newblock {\em Trends in Cognitive Sciences}.

\bibitem[Barron et~al., 2020]{barron2020prediction}
Barron, H.~C., Auksztulewicz, R., and Friston, K. (2020).
\newblock Prediction and memory: A predictive coding account.
\newblock {\em Progress in Neurobiology}, 192:101821.

\bibitem[Buhry et~al., 2011]{buhry2011reactivation}
Buhry, L., Azizi, A.~H., Cheng, S., et~al. (2011).
\newblock Reactivation, replay, and preplay: how it might all fit together.
\newblock {\em Neural Plasticity}.

\bibitem[Dalm et~al., 2023]{dalm2023effective}
Dalm, S., van Gerven, M., and Ahmad, N. (2023).
\newblock Effective learning with node perturbation in deep neural networks.
\newblock {\em arXiv preprint arXiv:2310.00965}.

\bibitem[Demircigil et~al., 2017]{demircigil2017model}
Demircigil, M., Heusel, J., L{\"o}we, M., Upgang, S., and Vermet, F. (2017).
\newblock On a model of associative memory with huge storage capacity.
\newblock {\em Journal of Statistical Physics}, 168:288--299.

\bibitem[Fischer and Igel, 2012]{fischer2012introduction}
Fischer, A. and Igel, C. (2012).
\newblock An introduction to restricted boltzmann machines.
\newblock {\em Progress in Pattern Recognition, Image Analysis, Computer Vision, and Applications}.

\bibitem[Friston, 2010]{friston2010free}
Friston, K. (2010).
\newblock The free-energy principle: a unified brain theory?
\newblock {\em Nature Reviews Neuroscience}, 11(2):127--138.

\bibitem[Friston et~al., 2006]{friston2006free}
Friston, K., Kilner, J., and Harrison, L. (2006).
\newblock A free energy principle for the brain.
\newblock {\em Journal of Physiology}, 100(1-3):70--87.

\bibitem[Graves et~al., 2016]{graves2016hybrid}
Graves, A., Wayne, G., Reynolds, M., Harley, T., Danihelka, I., Grabska-Barwi{\'n}ska, A., Colmenarejo, S.~G., Grefenstette, E., Ramalho, T., Agapiou, J., et~al. (2016).
\newblock Hybrid computing using a neural network with dynamic external memory.
\newblock {\em Nature}, 538(7626):471--476.

\bibitem[G{\"u}{\c{c}}l{\"u} and van Gerven, 2015]{gucclu2015deep}
G{\"u}{\c{c}}l{\"u}, U. and van Gerven, M. A.~J. (2015).
\newblock Deep neural networks reveal a gradient in the complexity of neural representations across the ventral stream.
\newblock {\em Journal of Neuroscience}, 35(27):10005--10014.

\bibitem[Guo et~al., 2020]{guo2020improved}
Guo, Y., Liu, M., Yang, T., and Rosing, T. (2020).
\newblock Improved schemes for episodic memory-based lifelong learning.
\newblock {\em Advances in Neural Information Processing Systems}.

\bibitem[Gupta et~al., 2018]{gupta2018meta}
Gupta, A., Mendonca, R., Liu, Y., Abbeel, P., and Levine, S. (2018).
\newblock Meta-reinforcement learning of structured exploration strategies.
\newblock {\em Advances in Neural Information Processing Systems}, 31.

\bibitem[Hancock and Kittler, 1993]{hancock1993bayesian}
Hancock, E.~R. and Kittler, J. (1993).
\newblock A bayesian interpretation for the hopfield network.
\newblock {\em IEEE International Conference on Neural Networks}.

\bibitem[Hasselmo, 2000]{hasselmo2000septal}
Hasselmo, M.~E. (2000).
\newblock Septal modulation of hippocampal dynamics: what is the function of the theta rhythm?
\newblock In {\em The behavioral neuroscience of the septal region}, pages 92--114. Springer.

\bibitem[Hasselmo and Cekic, 1996]{hasselmo1996suppression}
Hasselmo, M.~E. and Cekic, M. (1996).
\newblock Suppression of synaptic transmission may allow combination of associative feedback and self-organizing feedforward connections in the neocortex.
\newblock {\em Behavioural brain research}, 79(1-2):153--161.

\bibitem[Hasselmo et~al., 1997]{hasselmo1997noradrenergic}
Hasselmo, M.~E., Linster, C., Patil, M., Ma, D., and Cekic, M. (1997).
\newblock Noradrenergic suppression of synaptic transmission may influence cortical signal-to-noise ratio.
\newblock {\em Journal of Neurophysiology}, 77(6):3326--3339.

\bibitem[Hemmer and Steyvers, 2009]{hemmer2009bayesian}
Hemmer, P. and Steyvers, M. (2009).
\newblock A bayesian account of reconstructive memory.
\newblock {\em Topics in Cognitive Science}, 1(1):189--202.

\bibitem[Ho et~al., 2020]{ho2020denoising}
Ho, J., Jain, A., and Abbeel, P. (2020).
\newblock Denoising diffusion probabilistic models.
\newblock {\em Advances in Neural Information Processing Systems}.

\bibitem[Hopfield, 1982]{hopfield1982neural}
Hopfield, J.~J. (1982).
\newblock Neural networks and physical systems with emergent collective computational abilities.
\newblock {\em Proceedings of the National Academy of Sciences}, 79(8):2554--2558.

\bibitem[Jones and Wilkinson, 2020]{jones2020prediction}
Jones, M. and Wilkinson, S. (2020).
\newblock {\em From prediction to imagination}.
\newblock Cambridge University Press.

\bibitem[Kadam and Vaidya, 2020]{kadam2020review}
Kadam, S. and Vaidya, V. (2020).
\newblock Review and analysis of zero, one and few shot learning approaches.
\newblock {\em International Conference on Intelligent Systems Design and Applications}.

\bibitem[Kloeden et~al., 1992]{kloeden1992stochastic}
Kloeden, P.~E., Platen, E., Kloeden, P.~E., and Platen, E. (1992).
\newblock {\em Stochastic differential equations}.
\newblock Springer.

\bibitem[Kolodner, 1983]{kolodner1983reconstructive}
Kolodner, J.~L. (1983).
\newblock Reconstructive memory: A computer model.
\newblock {\em Cognitive Science}, 7(4):281--328.

\bibitem[Krotov, 2023]{krotov2023new}
Krotov, D. (2023).
\newblock A new frontier for hopfield networks.
\newblock {\em Nature Reviews Physics}, pages 1--2.

\bibitem[Krotov and Hopfield, 2021]{krotov2020large}
Krotov, D. and Hopfield, J. (2021).
\newblock Large associative memory problem in neurobiology and machine learning.
\newblock {\em International Conference on Learning Representations}.

\bibitem[Krotov and Hopfield, 2016]{krotov2016dense}
Krotov, D. and Hopfield, J.~J. (2016).
\newblock Dense associative memory for pattern recognition.
\newblock {\em Advances in Neural Information Processing Systems}.

\bibitem[Lopez-Paz and Ranzato, 2017]{lopez2017gradient}
Lopez-Paz, D. and Ranzato, M. (2017).
\newblock Gradient episodic memory for continual learning.
\newblock {\em Advances in Neural Information Processing Systems}, 30.

\bibitem[Michel and Farrell, 1990]{michel1990associative}
Michel, A.~N. and Farrell, J.~A. (1990).
\newblock Associative memories via artificial neural networks.
\newblock {\em IEEE Control Systems Magazine}, 10(3):6--17.

\bibitem[Millidge et~al., 2021]{millidge2021predictive}
Millidge, B., Seth, A., and Buckley, C.~L. (2021).
\newblock Predictive coding: a theoretical and experimental review.
\newblock {\em arXiv preprint arXiv:2107.12979}.

\bibitem[Nash et~al., 2015]{nash2015persuadability}
Nash, R.~A., W., R.~L., and Hope, L. (2015).
\newblock On the persuadability of memory: Is changing people's memories no more than changing their minds?
\newblock {\em British Journal of Psychology}, 106(2):308--326.

\bibitem[{\'O}lafsd{\'o}ttir et~al., 2018]{olafsdottir2018role}
{\'O}lafsd{\'o}ttir, H.~F., Bush, D., and Caswell, B. (2018).
\newblock The role of hippocampal replay in memory and planning.
\newblock {\em Current Biology}, 28(1):R37--R50.

\bibitem[Ororbia and Kifer, 2022]{ororbia2022neural}
Ororbia, A. and Kifer, D. (2022).
\newblock The neural coding framework for learning generative models.
\newblock {\em Nature communications}, 13(1):2064.

\bibitem[Ramsauer et~al., 2021]{ramsauer2020hopfield}
Ramsauer, H., Sch{\"a}fl, B., Lehner, J., Seidl, P., Widrich, M., Adler, T., Gruber, L., Holzleitner, M., Pavlovi{\'c}, M., Sandve, G.~K., et~al. (2021).
\newblock Hopfield networks is all you need.
\newblock {\em Internetional Conference on Learning Representations}.

\bibitem[Rao and Ballard, 1999]{rao1999predictive}
Rao, R. P.~N. and Ballard, D.~H. (1999).
\newblock Predictive coding in the visual cortex: a functional interpretation of some extra-classical receptive-field effects.
\newblock {\em Nature Neuroscience}, 2(1):79--87.

\bibitem[Raya and Ambrogioni, 2023]{raya2023spontaneous}
Raya, G. and Ambrogioni, L. (2023).
\newblock Spontaneous symmetry breaking in generative diffusion models.
\newblock {\em arXiv preprint arXiv:2305.19693}.

\bibitem[Roediger and McDermott, 1995]{roediger1995creating}
Roediger, H.~L. and McDermott, K.~B. (1995).
\newblock Creating false memories: Remembering words not presented in lists.
\newblock {\em Journal of Experimental Psychology: Learning, Memory, and Cognition}, 21(4):803.

\bibitem[S. et~al., 2021]{song2021scorebased}
S., Y., S., J., K., D.~P., K., A., E., S., and P., B. (2021).
\newblock Score-based generative modeling through stochastic differential equations.
\newblock {\em International Conference on Learning Representations}.

\bibitem[Salvatori et~al., 2021]{salvatori2021associative}
Salvatori, T., Song, Y., Hong, Y., Sha, L., Frieder, S., Xu, Z., Bogacz, R., and Lukasiewicz, T. (2021).
\newblock Associative memories via predictive coding.
\newblock {\em Advances in Neural Information Processing Systems}, 34:3874--3886.

\bibitem[Schacter et~al., 2007]{schacter2007remembering}
Schacter, D.~L., Addis, D.~R., and B., R.~L. (2007).
\newblock Remembering the past to imagine the future: the prospective brain.
\newblock {\em Nature Reviews Neuroscience}, 8(9):657--661.

\bibitem[Sejnowski and Tesauro, 1989]{sejnowski1989hebb}
Sejnowski, T.~J. and Tesauro, G. (1989).
\newblock The hebb rule for synaptic plasticity: algorithms and implementations.
\newblock In {\em Neural Models of Plasticity}, pages 94--103. Elsevier.

\bibitem[Sohl-Dickstein et~al., 2015]{sohl2015deep}
Sohl-Dickstein, J., Weiss, E., Maheswaranathan, N., and Ganguli, S. (2015).
\newblock Deep unsupervised learning using nonequilibrium thermodynamics.
\newblock {\em International Conference on Machine Learning}.

\bibitem[Song et~al., 2021]{song2020denoising}
Song, J., Meng, C., and Ermon, S. (2021).
\newblock Denoising diffusion implicit models.
\newblock {\em Internetional Conference on Learning Representations}.

\bibitem[Spratling, 2017]{spratling2017review}
Spratling, M.~W. (2017).
\newblock A review of predictive coding algorithms.
\newblock {\em Brain and Cognition}, 112:92--97.

\bibitem[Stachenfeld et~al., 2017]{stachenfeld2017hippocampus}
Stachenfeld, K.~L., Botvinick, M.~M., and Gershman, S.~J. (2017).
\newblock The hippocampus as a predictive map.
\newblock {\em Nature Neuroscience}, 20(11):1643--1653.

\end{thebibliography}

\appendix 

\section{Details of the experiments} \label{supp: experimental details}
Binary patterns were generated independently by applying the sign function to standard Gaussian vectors. To test recovery, we corrupted the patterns using the formula $\tilde{\vect{y}} = \theta \vect{y} + \sqrt{1 - \theta^2} \vect{\delta}$, where $\vect{\delta}$ is a standard Gaussian vector. In all the experiments, we used $\theta = 0.68$. 

We compared the storage and recovery performance of several models. Modern Hopfield networks were implemented with the iterative update given in \citep{ramsauer2020hopfield}. We set $\beta$ to be equal to $5$ and we applied $150$ iterations to each noisy pattern. 

For reasons of stability, we used variance-preserving diffusion models defined by the following noise-injection SDE:
\begin{equation}
    \text{d} \vect{x}_t = - \gamma \vect{x}_t + \gamma \text{d} W_t~,
\end{equation}
where $W_t$ is a standard Wiener process. This is slighly different from the variance-exploding models discussed in the main text, but it delivers nearly identical results. The diffusion models were used as denoiser by applying the deterministic ODE dynamics given in \citep{song2021scorebased}: 
\begin{equation}
    \frac{\text{d} \vect{x}_t}{\text{d} t} = \frac{1}{2} \left(\vect{x}_t + \sigma^2 \nabla_{\vect{x}} \log{p_{t}(\vect{x}(t))} \right)~,
\end{equation}
which exactly reproduces the marginal densities of the stochastic dynamics. We integrated the dynamics using $300$ stepos of Euler integration. The initial time was set to match the match level using the following formula: $t_{\text{start}} = - \gamma^{-1} 2 \log(\theta)$, with $\gamma = 0.8$. This formula implies the use of a constant noise scheduling with variance equal to $\gamma$.

In exact score-based diffusion model, we used the exact formula for the score given the marginal density:
\begin{equation}
    \nabla \log{p_t(\vect{x})} = (\vect{x} - Y \vect{h}_t(\vect{x}))/(1 - \theta_t^2)~,
\end{equation}
where $Y$ is a matrix having the patterns $\vect{y}^j$ as columns. The weight vector $\vect{h}_t$ is obtained using the softmax function:
\begin{equation}
    h_{t,j}(\vect{x}) = \text{softmax}\left(\dots, -\norm{\vect{x} - \vect{y}^k}{2}^2 /(2 (1 - \theta_t^2), \dots \right)_j~,
\end{equation}
which depends on the correlation between the state $\vect{x}$ and each of the patterns. This is very similar to the softmax formula given in the update of modern Hopfield networks, which is unsurprising given the equivalence of their energy functions. 

Learned generative diffusion models used a three layers fully connected architecture with reLu non-linearities, $d$ input and output units and $80 d$ hidden units in each layer. The time index was embedded by converting it to $\theta_t = \exp(-0.5 \gamma t)$ and then by concatenating this value to each layer. They were trained using the optimizer Adam with base rate $0.001$ and fixed batches containing all the generated patterns. 

For the Hopfield and exact models, the capacity was estimated by evaluating the average reconstruction error in a grid of dimensionalities and number of patterns. Given each combination, the error was computed $140$ using randomly sampled patterns. The capacity was then estimated by finding the maximum number of patterns such that, for a given dimensionality, the average Hamming error did not exceed a threshold value of $3 \%$. Error intervals were obtained by bootstrap re-sampling of the error. In order to increase the snd, we smoother both the mean capacity and the bounds by convolving them with a Gaussian kernel with $\sigma = 1.5$.

This estimation method would have been too expensive for evaluating the capacity of learned models, since the network needs to be fully re-trained for any given set of patterns. 

Instead, for a given dimensionality $d$, we trained the network with $n_d$ patterns, where $n_d$ is the estimated capacity of the exact score model plus $4$. After training we evaluated the error on $30$ batches of unseen noise corrupted versions of the training patterns. If the error was below threshold, we returned $n_d$ as the estimated capacity, otherwise, we reduced $n_d$ by one and repeated the treaning until the error was below threshold. This procedure was repeated $8$ times for $d$ equal to $10$, $12$, $14$ and $16$.

\end{document}